\documentclass{article}
\usepackage{ijcai19}

\usepackage{verbatim}
\usepackage{times}
\usepackage{soul}
\usepackage{url}
\usepackage[hidelinks]{hyperref}
\usepackage[utf8]{inputenc}
\usepackage[small]{caption}
\usepackage{graphicx}
\usepackage{amsmath}
\usepackage{booktabs}
\usepackage{algorithm}
\usepackage{algorithmic}
\hypersetup{draft}

\title{Graph Convolutional Networks using Heat Kernel for Semi-supervised Learning}

\author{
Bingbing Xu\and
Huawei Shen\and
Qi Cao\and
Keting Cen\And
Xueqi Cheng
\affiliations
CAS Key Laboratory of Network Data Science and Technology, \\Institute of Computing Technology, Chinese Academy of Sciences\\
School of Computer Science and Technology,\\
University of Chinese Academy of Sciences,
Beijing, China\\
\emails
\{xubingbing, shenhuawei, caoqi, cenketing, cxq\}@ict.ac.cn
}

\begin{document}

\maketitle

\begin{abstract}
Graph convolutional networks gain remarkable success in semi-supervised learning on graph-structured data. The key to graph-based semi-supervised learning is capturing the smoothness of labels or features over nodes exerted by graph structure. Previous methods, spectral methods and spatial methods, devote to defining graph convolution as a weighted average over neighboring nodes, and then learn graph convolution kernels to leverage the smoothness to improve the performance of graph-based semi-supervised learning. One open challenge is how to determine appropriate neighborhood that reflects relevant information of smoothness manifested in graph structure. In this paper, we propose GraphHeat, leveraging heat kernel to enhance low-frequency filters and enforce smoothness in the signal variation on the graph. GraphHeat leverages the local structure of target node under heat diffusion to determine its neighboring nodes flexibly, without the constraint of order suffered by previous methods. GraphHeat achieves state-of-the-art results in the task of graph-based semi-supervised classification across three benchmark datasets: Cora, Citeseer and Pubmed.
\end{abstract}

\section{Introduction}
\label{introduction}

Convolutional neural networks (CNNs)~\cite{lecun1998gradient} have been successfully used in various machine learning problems, such as image classification~\cite{he2016deep} and speech recognition~\cite{hinton2012deep}, where there is an underlying Euclidean structure. However, in many research areas, data are naturally located in a non-Euclidean space, with graph or network being one typical case. The success of CNNs motivates researchers to design convolutional neural network on graphs. Existing methods fall into two categories, spatial methods and spectral methods, according to the way that convolution is defined. 

Spatial methods define convolution directly in the vertex domain, following the practice of the conventional CNN. For each node, convolution is defined as a weighted average function over its neighboring nodes, with the weighting function characterizing the influence exerting to the target node by its neighboring nodes. GraphSAGE~\cite{hamilton2017inductive} defines the weighting function as various aggregators over neighboring nodes. Graph attention network (GAT) proposes learning the weighting function via self-attention mechanism~\cite{velickovic2017graph}. MoNet~\cite{monti2017geometric} offers us a general framework for designing spatial methods. It takes convolution as a weighted average of multiple weighting functions defined over neighboring nodes. One open challenge for spatial methods is how to determine appropriate neighborhood for target node when defining graph convolution. 

Spectral methods define graph convolution via convolution theorem. As the pioneering work of spectral methods, Spectral CNN~\cite{bruna2014spectral} leverages graph Fourier transform to convert signals defined in vertex domain into spectral domain, and defines the convolution kernel as a set of learnable coefficients associated with Fourier bases, i.e., the eigenvectors of Laplacian matrix. Given that the magnitude of eigenvalue reflects the smoothness over graph of the associated eigenvector, only the eigenvectors associated with smaller eigenvalues are used. Unfortunately, this method relies on the eigendecomposition of Laplacian matrix, resulting in high computational complexity. ChebyNet~\cite{defferrard2016convolutional} introduces a polynomial parametrization to convolution kernel, i.e., convolution kernel is taken as a polynomial function of the diagonal matrix of eigenvalues. Subsequently, Kipf and Welling~\cite{kipf2017semi} proposed graph convolutional network (GCN) via a localized first-order approximation to ChebyNet. However, both ChebyNet and GCN fail to filter out high-frequency noise carried by eigenvectors associated with high eigenvalues. GWNN~\cite{xu2018graph} leverages graph wavelet to implement localized convolution. In sum, to the best of our knowledge, previous works lack an effective graph convolution method to capture the smoothness manifested in the structure of network.

\label{strengths}
In this paper, we propose graph convolutional network with heat kernel, namely GraphHeat, for graph-based semi-supervised learning. Different from existing spectral methods, GraphHeat uses heat kernel to assign larger importance to low-frequency filters, explicitly discounting the effect of high-frequency variation of signals on graph. In this way, GraphHeat performs well at capturing the smoothness of labels or features over nodes exerted by graph structure. From the perspective of spatial methods, GraphHeat leverages the process of heat diffusion to determine neighboring nodes that reflect the local structure of the target node and the relevant information of smoothness manifested in graph structure. Experiments show that GraphHeat achieves state-of-the-art results in the task of graph-based semi-supervised classification across benchmark datasets: Cora, Citeseer and Pubmed.

\section{Preliminary}

\subsection{Graph Definition}
$G=\{V,E,A\}$ denotes an undirected graph, where $V$ is the set of nodes with $|V|=n$, $E$ is the set of edges, and $A$ is the adjacency matrix with $A_{i,j} = A_{j,i}$ to define the connection between node $i$ and node $j$. Graph Laplacian matrix is defined as $\mathcal{L}=D-A$ where $D$ is a diagonal degree matrix with $D_{i,i}=\sum_{j}A_{i,j}$, normalized Laplacian matrix $L= I_{n}-D^{-1/2}AD^{-1/2}$ where $I_{n}$ is the identity matrix. Since $L$ is a real symmetric matrix, it has a complete set of orthonormal eigenvectors $U=(u_1,u_2,...,u_{n})$, known as Laplacian eigenvectors. These eigenvectors have associated real, non-negative eigenvalues $\{\lambda_l\}_{l=1}^{n}$, identified as the frequencies of the graph. Without loss of generality, we have $\lambda_1\leq \lambda_2 \leq \cdots \leq \lambda_n$. Using eigenvectors and eigenvalues of $L$, we have $L = U\Lambda U^\top$, where $\Lambda =$diag$(\{\lambda_l\}_{l=1}^{n})$.

\subsection{Graph Fourier Transform}
Taking the eigenvectors of normalized Laplacian matrix as a set of bases, graph Fourier transform of a signal $x\in R^n$ on graph $G$ is defined as $\hat x=U^\top x$, and the inverse graph Fourier transform is $x=U\hat x$~\cite{shuman2013emerging}. According to convolution theorem, graph Fourier transform offers us a way to define the graph convolution operator, represented as $*_{G}$.  $f$ denotes the convolution kernel in spatial domain, and $*_{G}$ is defined as

\vspace{0mm}
\begin{equation}
x *_{G} f = U\big((U^\top f)\odot(U^\top x)\big),
\label{eq1}
\vspace{0mm}
\end{equation}
where $\odot$ is the element-wise Hadamard product.

\subsection{Graph Convolutional Networks}
\label{wpmodel}
With the vector $U^\top f$ replaced by a diagonal matrix $g_\theta$, Hadamard product can be written in the form of matrix multiplication. Filtering the signal $x$ by convolution kernel $g_\theta$, Spectral CNN~\cite{bruna2014spectral} is obtained as 

\vspace{0mm}
\begin{equation}
y=Ug_\theta U^\top x=(\theta_1u_1u_1^\top+\theta_2u_2u_2^\top+\cdots+\theta_nu_nu_n^\top)x,
\label{eq:spec}
\vspace{0mm}
\end{equation}
where $g_\theta=$diag$(\{\theta_i\}_{i=1}^n)$ is defined in spectral domain.

The above non-parametric convolution kernel is not localized in space and its parameter complexity scales up to $O(n)$. To combat these issues, ChebyNet~\cite{defferrard2016convolutional} is proposed to parameterize $g_\theta$ with a polynomial expansion

\begin{equation}
g_\theta=\sum_{k=0}^{K-1}\alpha_k \Lambda^k,
\label{eq21}
\vspace{0mm}
\end{equation}
where $K$ is a hyper-parameter and the parameter $\alpha_k$ is the polynomial coefficient. Then graph convolution operation in ChebyNet is defined as
\begin{equation}
\begin{aligned}
y&=Ug_\theta U^\top x=U (\sum_{k=0}^{K-1}\alpha_k \Lambda^k) U^\top x\\&=\{\alpha_0I+\alpha_1(\lambda_1u_1u_1^\top+\cdots+\lambda_nu_nu_n^\top)\\&+\cdots+\alpha_{K-1}(\lambda_1^{K-1}u_1u_1^\top+\cdots+\lambda_n^{K-1}u_nu_n^\top)\}x\\&=(\alpha_0I+\alpha_1 L+\alpha_2 L^2+\cdots +\alpha_{K-1} L^{K-1})x.
\end{aligned}
\label{eq:cheby}
\vspace{0mm}
\end{equation}


\vspace{0mm}
GCN~\cite{kipf2017semi} simplifies ChebyNet by only considering the first-order polynomial approximation, i.e., $K=2$, and setting $\alpha = \alpha_0=-\alpha_1$. Graph convolution is defined as

\vspace{0mm}
\begin{equation}
\begin{aligned}
y&=Ug_\theta U^\top x=U (\sum_{k=0}^{1}\alpha_k \Lambda^k) U^\top x\\
&=\{\alpha I-\alpha(\lambda_1u_1u_1^\top+\cdots+\lambda_nu_nu_n^\top)\}x\\
& =\alpha( I-L)x.
\end{aligned}
\label{eq:gcn}
\vspace{0mm}
\end{equation}

Note that the above three methods all take $\{u_iu_i^\top\}_{i=1}^n$ as a set of basic filters. Spectral CNN directly learns the coefficients of each filter. ChebyNet and GCN parameterize the convolution kernel $g_\theta$ to get combined filters, e.g., $L$, accelerating the computation and reducing parameter complexity.

\section{GraphHeat for Semi-supervised Learning}

Graph smoothness is the prior information over graph that connected nodes tend to have the same label or similar features. Graph-based semi-supervised learning, e.g., node classification, gains success via leveraging the smoothness of labels or features over nodes exerted by graph structure. Graph convolutional neural network offers us a promising and flexible framework for graph-based semi-supervised learning. In this section, we analyze the weakness of previous graph convolution neural networks at capturing smoothness manifested in graph structure, and propose GraphHeat to circumvent the problem of graph-based semi-supervised learning.


\subsection{Motivation}

Given a signal $x$ defined on graph, its smoothness with respect to the graph is measured using

\begin{equation}
x^\top L x = \sum_{(a,b)\in E}A_{a,b}{\big( \frac{x(a)}{\sqrt{d_a}}-\frac{x(b)}{\sqrt{d_b}}\big)}^2,
\label{eq:xsmooth}
\end{equation}
where $a,b \in V$, $x(a)$ represents the value of signal $x$ on node $a$, $d_a$ is the degree of node $a$. The smoothness of signal $x$ characterizes how likely it has similar values, normalized by degree, on connected nodes.


For an eigenvector $u_i$ of the normalized Laplacian matrix $L$, its associated eigenvalue $\lambda_i$ captures the smoothness of $u_i$~\cite{shuman2013emerging} as

\begin{equation}
\lambda_i = u_i^\top L u_i.
\label{eq:smooth}
\end{equation}

According to Eq.~(\ref{eq:smooth}), eigenvectors associated with small eigenvalues are smooth with respect to graph structure, i.e., they carry low-frequency variation of signals on graph. In contrast, eigenvectors associated with large eigenvalues are signals with high-frequency, i.e., less smoother signals.   


Note that the eigenvectors $U=(u_1,u_2,\cdots,u_n)$ are orthonormal, i.e., 
\begin{equation}
\begin{aligned}
&u_i^\top u_i =1 \  \ \ \ \rm{and}
&u_i^\top u_j =0, \ \ j\neq i.
\label{eq:orgh}
\end{aligned}
\end{equation}

Given a signal $x$ defined on graph, we have
\begin{equation}
x=\alpha_1 u_1 + \alpha_2 u_2+\cdots+\alpha_n u_n,
\label{eq:rep}
\end{equation}
where $\{\alpha_i\}_{i=1}^n$ is the coefficient of each eigenvector. Each eigenvector $u_i$ offers us a basic filter, i.e., $u_iu_i^T$. Only the component $\alpha_iu_i$ of $x$ can pass the filter:
\begin{equation}
u_i u_i^\top x=\alpha_1u_i u_i^\top u_1 +\cdots+ \alpha_n u_iu_i^\top u_n=\alpha_i u_i.
\label{eq:fil}
\end{equation}
Therefore, $\{u_iu_i^\top\}_{i=1}^n$ form a set of basic filters, and the eigenvalue associated with $u_i$ represents the frequency of signals that can pass the filter $u_iu_i^\top$, i.e., $\alpha_i u_i$.

We now use these basic filters to analyze the weakness of previous methods at capturing the smoothness manifested in graph structure. Spectral CNN, Eq.~(\ref{eq:spec}), directly learns the coefficients of each basic filter to implement graph convolution. ChebyNet, Eq.~(\ref{eq:cheby}), defines graph convolution based on a set of combined filters, i.e., $\{L^{k}\}_{k=0}^{K-1}$, which are obtained through assigning higher weights to high-frequency basic filters. For example, for a combined filter $L^k$, the weight assigned to $u_iu_i^T$ is $\lambda_i^k$, which increases with respect to $\lambda_i$. GCN only uses the first order approximation, i.e., $L$, for semi-supervised learning. In sum, the three methods do not suppress high-frequency signals by assigning lower weight to high-frequency basic filters, failing to or not well capture the smoothness manifested in graph structure.

\subsection{GraphHeat: Graph Convolutional using Heat Kernel}

We propose GraphHeat to capture the smoothness of labels or features over nodes exerted by graph structure. GraphHeat comprises a set of combined filters which discount high-frequency basic filters via heat kernel (see~\cite{chung1997spectral}, Chapter10). 

Heat kernel is defined as 
\begin{equation}
f(\lambda_i)=e^{-s\lambda_i},
\label{eq:hk}
\vspace{0mm}
\end{equation}
where $s\geq 0$ is a scaling hyper-parameter. For clarity, we denote with $\Lambda_s$ as $\Lambda_s=$diag$(\{e^{-s \lambda_i}\}_{i=1}^n)$. Using heat kernel, we define the convolution kernel  as


\begin{equation}
g_\theta=\sum_{k=0}^{K-1}\theta_k (\Lambda_s)^k,
\label{eq3}
\vspace{0mm}
\end{equation}
where $\theta_k$ is the parameter. 

For a signal $x$, graph convolution is achieved by

\begin{equation}
\begin{aligned}
y&=Ug_\theta U^\top x=U (\sum_{k=0}^{K-1}\theta_k (\Lambda_s)^k) U^\top x\\&=\{\theta_0I+\theta_1(e^{-s\lambda_1}u_1u_1^\top+\cdots+e^{-s\lambda_n}u_nu_n^\top)+\cdots\\&+\theta_{K-1}(e^{-(K-1)s\lambda_1}u_1u_1^\top+\cdots+ e^{-({K-1})s\lambda_n} u_nu_n^\top)\}x\\
&=(\theta_0I+\theta_1 e^{-sL}+\theta_2 e^{-2sL}+\cdots +\theta_{K-1} e^{-(K-1)sL})x.
\end{aligned}
\label{eq:gh}
\vspace{0mm}
\end{equation}

The key insight of GraphHeat is that it achieves a smooth graph convolution via discounting high-frequency basic filters. The weight assigned to the basic filter $u_iu_i^\top$ is $e^{-ks\lambda_i}$, which decreases with respect to $\lambda_i$. This distinguishes GraphHeat from existing methods that promotes high-frequency basic filters.



To reduce parameter complexity for semi-supervised learning, we only retain the first two items in Eq.~(\ref{eq:gh})

\begin{equation}
y=(\theta_0I+\theta_1 e^{-sL})x.
\label{eq4}
\vspace{0mm}
\end{equation}

The computation of $e^{-sL}$ is completed via Chebyshev polynomials without eigendecomposition of $L$~\cite{hammond2011wavelets}. The computational complexity is $O(m\times|E|)$, where $|E|$ is the number of edges, and $m$ is the order of Chebyshev polynomials. Such a linear complexity makes GraphHeat applicable to large-scale networks. 

\begin{figure}
\centering
\includegraphics[width=.45\textwidth]{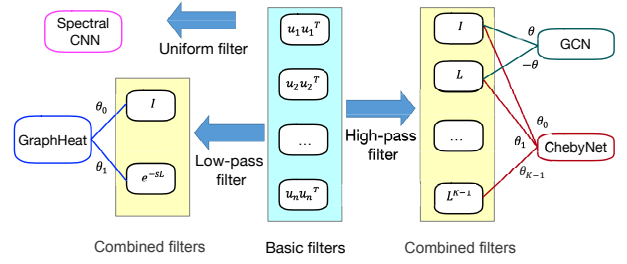}
\caption{The connection and difference between GraphHeat and  previous graph convolution methods. GraphHeat captures the smoothness over graph by suppressing high-frequency signals, acting like a low-pass filter. In contrast, previous methods can be viewed as high-pass filters or uniform filters.}
\label{fig:pmodel}
\end{figure}

Figure~\ref{fig:pmodel} illustrates the connection and difference between our GraphHeat and  previous graph convolution methods. GraphHeat captures the smoothness over graph by suppressing high-frequency signals, acting like a low-pass filter. In contrast, previous methods can be viewed as high-pass filters or uniform filters, lacking the desirable capability to filter out high-frequency variations over graph. Therefore, GraphHeat is expected to perform better on graph-based semi-supervised learning. 


\subsection{GraphHeat: Defining Neighboring Nodes under Heat Diffusion}

In the previous subsection, GraphHeat is formulated as a kind of spectral method for graph convolution. The benefit of GraphHeat is also highlighted through comparing it with existing spectral methods from the perspective of frequency and graph signal processing. Now we offer an understanding of GraphHeat by taking it as a kind of spatial method.

Spatial methods define graph convolution as a weighted average over the neighboring nodes of target node. A general framework for spatial methods is to define graph convolution as a weighted average of a set of weighting functions, with each weighting function characterizing certain influence exerting to the target node by its neighboring nodes~\cite{monti2017geometric}. For GraphHeat, the weighting functions are defined via heat kernels, i.e., $e^{-ksL}$, and graph convolution kernel is the coefficients $\theta_k$. 

In GraphHeat, each $e^{-ksL}$ actually corresponds to a similarity metric among nodes under heat diffusion. The similarity between node $i$ and node $j$ characterizes the amount of energy received by node $j$ when a unit of heat flux or energy is offered to node $i$, and vice versa. Indeed, the scaling parameter $s$ acts like the length of time during which diffusion proceeds, and $k$ represents different energy levels.

With the heat diffusion similarity, GraphHeat offers us a way to define neighboring nodes for target node. Specifically, for a target node, nodes with the similarity higher than a threshold $\epsilon$ are regarded as neighboring nodes. Such a way of defining neighboring nodes is fundamentally different from previous graph convolution methods, which usually adopts an order-style way, i.e., neighboring nodes are within $K$-hops away from target node.

\begin{figure}
\centering
\includegraphics[width=.45\textwidth]{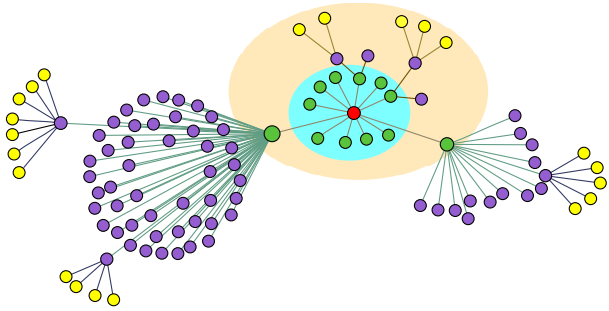}
\caption{Neighboring nodes defined by GraphHeat and previous methods that are based on shortest-path distance. Target node is colored in red, and the 1-st, 2-nd, 3-rd order neighboring nodes are colored in green, purple, and yellow, respectively. The small circle marks neighboring nodes of GraphHeat with a small $s$. The big circle includes neighboring nodes when $s$ is larger.}
\label{fig:scalefreenetwork}
\end{figure}

Figure~\ref{fig:scalefreenetwork} illustrates the difference between the two kinds of ways for defining neighboring nodes. Previous methods define neighboring nodes according to the shortest path distance, $K$, away from the target node (the red node in Figure~\ref{fig:scalefreenetwork}). When $K=1$, only green nodes are included, and this may lead to ignoring some relevant nodes. When $K=2$, all purple neighbors are included. This may bring noise to the target node, since connections with high-degree nodes may represent popularity of high-degree nodes instead of correlation. Compared with constraining neighboring nodes via the shortest path distance, our method has the following benefits based on heat diffusion: (1) GraphHeat defines neighboring nodes in a continuous manner, via tuning the scaling parameter $s$; (2) It is flexible to leverage high-order neighbors while discarding some irrelevant low-order neighbors; (3) The range of neighboring nodes varies across target node, which is demonstrated in Section~\ref{Experiments:Case}; (4) Parameter complexity does not increase with the order of neighboring nodes, since $e^{-sL}$ can include the relation of all neighboring nodes in one matrix.

\begin{figure*}
\centering
\includegraphics[width=0.95\textwidth,height=4.5cm]{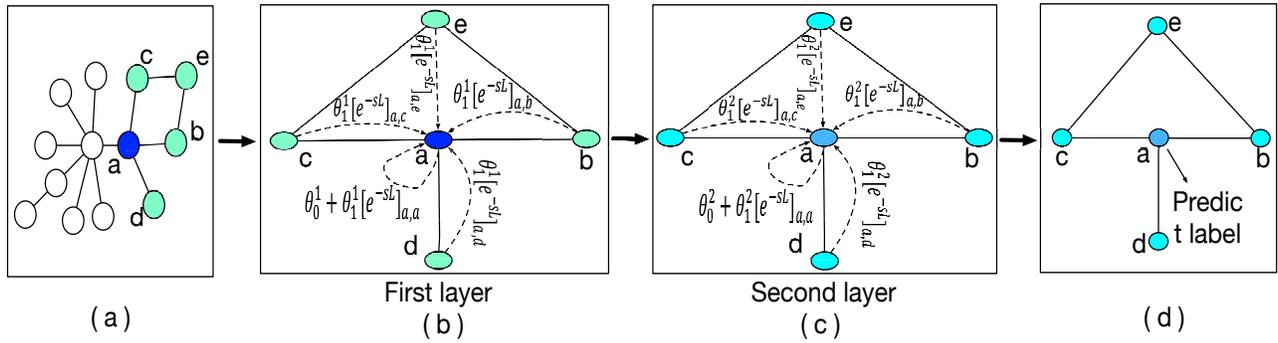}
\caption{Two-layer GraphHeat for semi-supervised node classication. (a) select neighboring nodes (solid nodes) for target node $a$ via $e^{-sL}$. (b-c) the solid line represents edge in graph. Dotted line represents the weights between selected nodes and target node $a$. Representation of $a$ is updated through weighted average of solid nodes and itself. Node color reflects the update of node representation. (d) label prediction using updated representation in the last layer.}
\label{fig:architecture}
\end{figure*}

\subsection{Architecture}

Semi-supervised node classification assigns labels to unlabeled nodes according to the feature matrix of nodes and graph structure, supervised by the labels of a small set of nodes. In this paper, we consider a two-layer GraphHeat for semi-supervised node classification. The architecture of our model is
\begin{equation}
\begin{aligned}
&{\rm first  \ layer}:\  \\
&X^2_{j}={\rm ReLU}(\sum_{i=1}^{p} (\theta^1_{0,i,j}I+\theta^1_{1,i,j} e^{-sL}) X^1_{i}),
j=1,\cdots,q,&\\
&{\rm second  \ layer}:\  \\
&Z_j={\rm softmax}(\sum_{i=1}^{q} (\theta^2_{0,i,j}I+\theta^2_{1,i,j} e^{-sL}) X^2_{i}), j=1,\cdots,c,
\label{eqr2}
\end{aligned}
\end{equation}
where $p$ is the number of input feature, $q$ is the number of output feature in first layer, $X^m_{i}$ is the $i$-th feature of $m$-th layer, $\theta_{0,i,j}^m$ and $\theta_{1,i,j}^m$ are parameters in $m$-th layer, $c$ is the number of classes in node classification, $Z_j$ is an $n$-dimensional vector representing the prediction result for all nodes in class $j$. The loss function is the cross-entropy error over all labeled nodes:
\begin{equation}
Loss = -\sum_{i\in y_{L}} \sum_{j=1}^c Y_{ij}{\rm ln} Z_j{(i)},
\label{eqr1}
\end{equation}
where $y_{L}$ is the set of labeled nodes, $Y_{ij}=1$ if the label of node $i$ is $j$, and $Y_{ij}=0$ otherwise. The parameters $\theta$ are trained using gradient descent. Figure~\ref{fig:architecture} shows the architecture of our model.


\section{Experiments}

We evaluate the effectiveness of GraphHeat on three benchmarks. A detailed analysis about the influence of hyper-parameter is also conducted. Lastly, we show a case to intuitively demonstrate the strengths of our method.

\subsection{Datasets}

To evaluate the proposed method, we conduct experiments on three benchmark datasets, namely, Cora, Citeseer and Pubmed~\cite{sen2008collective}. In these citation network datasets, nodes represent documents and edges are citation links. Table~\ref{table:Datasetstatistics} shows an overview of three datasets. Label rate denotes the proportion of labeled nodes for training.

\begin{table}[htbp]
\centering
\vspace{0mm}
\small
\begin{tabular*}{\hsize}{@{}@{\extracolsep{\fill}}l r r r r r r@{}}
\hline
\textbf{Datasets} & \textbf{Nodes} & \textbf{Edges} & \textbf{Classes} & \textbf{Features} & \textbf{Label Rate}\\
\hline 
Cora & 2,708 & 5,429 & 7 & 1,433 & 0.052\\
Citeseer  & 3,327 & 4,732 & 6 & 3,703 & 0.036\\
Pubmed & 19,717 & 44,338 & 3 & 500 & 0.003\\
\hline
\end{tabular*}
\caption{The statistics of datasets}
\label{table:Datasetstatistics} 
\end{table}

\subsection{Baselines}

We compare with some traditional graph semi-supervised learning methods, including  MLP, only leveraging node features for classification, manifold regularization (ManiReg)~\cite{belkin2006manifold}, semi-supervised embedding (SemiEmb)~\cite{weston2012deep}, label propagation (LP)~\cite{zhu2003semi}, graph embeddings (DeepWalk)~\cite{perozzi2014deepwalk}, iterative classification algorithm (ICA)~\cite{lu2003link} and Planetoid~\cite{yang2016revisiting}. Furthermore, since graph convolutional networks are proved to be effective in semi-supervised learning, we also compare against the state-of-the-art spectral graph convolutional networks, i.e., ChebyNet~\cite{defferrard2016convolutional}, GCN~\cite{kipf2017semi}, and the state-of-the-art spatial convolutional networks, i.e., MoNet~\cite{monti2017geometric} and GAT~\cite{velickovic2017graph}.


\subsection{Experimental Settings}

We train a two-layer GraphHeat with 16 hidden units, and prediction accuracy is evaluated on a test set of 1000 labeled nodes. The partition of datasets is the same as GCN~\cite{kipf2017semi} with an additional validation set of 500 labeled samples to determine hyper-parameters.

Weights are initialized following~\cite{glorot2010understanding}. We adopt the Adam optimizer~\cite{kingma2014adam} for parameter optimization with an initial learning rate $lr = 0.01$. If $e^{-sL}$ is smaller than a given threshold $\epsilon$, we set it as zero to accelerate the computation and avoid noise. The optimal hyper-parameters, e.g., scaling parameter $s$ and threshold $\epsilon$, are chosen through validation set. For Cora, $s=3.5$ and $\epsilon=1e-4$. For Citeseer, $s=4.5$ and $\epsilon=1e-5$. For Pubmed, $s=3.0$ and $\epsilon=1e-5$. To avoid overfitting, dropout~\cite{srivastava2014dropout} is applied and the value is set as 0.5. The training process is terminated if the validation loss does not decrease for 200 consecutive epochs.

\subsection{Performance on Node Classification Task}

We now validate the effectiveness of  GraphHeat on node classification. Similar to previous methods, we use classification accuracy metric for quantitative evaluation. Experimental results are reported in Table~\ref{table:results}. Graph convolutional networks, including spatial methods and spectral methods, all perform much better than previous methods. This is due to that graph convolutional networks are trained in an end-to-end manner, and update representations via graph structure under the guide of labels. 


\begin{table}[htbp]
\small
\centering
\begin{tabular*}{\hsize}{@{}@{\extracolsep{\fill}}l l l l@{}}
\hline
\textbf{Method} & \textbf{Cora} & \textbf{Citeseer} & \textbf{Pubmed} \\
\hline
MLP & 55.1\% & 46.5\% & 71.4\% \\
ManiReg & 59.5\% &60.1\% & 70.7\% \\
SemiEmb & 59.0\% & 59.6\%& 71.7\% \\
LP & 68.0\% & 45.3\%& 63.0\% \\
DeepWalk & 67.2\% & 43.2\%& 65.3\% \\
ICA & 75.1\% & 69.1\%& 73.9\% \\
Planetoid & 75.7\% & 64.7\%& 77.2\% \\
\hline
ChebyNet & 81.2\% & 69.8\%& 74.4\% \\
GCN & 81.5\% & 70.3\%& 79.0\% \\
\hline
MoNet & 81.7$\pm$0.5\% & --- & 78.8$\pm$0.3\% \\
GAT & 83.0$\pm$0.7\% & \textbf{72.5$\pm$0.7\%} & 79.0$\pm$0.3\% \\
\hline
GraphHeat & \textbf{83.7\%} & \textbf{72.5\%} & \textbf{80.5\%} \\
\hline
\end{tabular*}
\caption{Results of node classification}
\label{table:results}
\end{table}



As for our GraphHeat model, it outperforms all baseline methods, achieving state-of-the-art results on all the three datasets. Note that the key to graph-based semi-supervised learning is to capture the smoothness
of labels or features over nodes exerted by graph structure. Different from all previous graph convolutional networks, our GraphHeat achieves such a smooth graph convolution via discounting high-frequency basic filters. In addition, we provide our neighboring nodes via heat diffusion and modulate parameter $s$ to suit diverse networks. In Cora and Pubmed, especially the Pubmed which has the smallest label rate, GraphHeat achieves its superiority significantly. In Citeseer which is sparser than the other two datasets, our method matches GAT. It may be due to that GAT learns the weight based on representation of nodes in hidden layer, while our model uses Laplacian matrix which only depends on the sparser graph.

\subsection{Influence of Hyper-parameter $s$ and $\epsilon$}
\label{sec:Influenceparameter} 
GraphHeat uses $e^{-sL}$ as the combined filter, where $s$ is a modulating parameter. As $s$ becomes larger, the range of feature diffusion becomes larger. Additionally, $e^{-sL}$ is truncated via $\epsilon$. This setting can speed up computation and remove noise.

\begin{figure}[htbp]
\centering
\includegraphics[width=.45\textwidth,height=6cm]{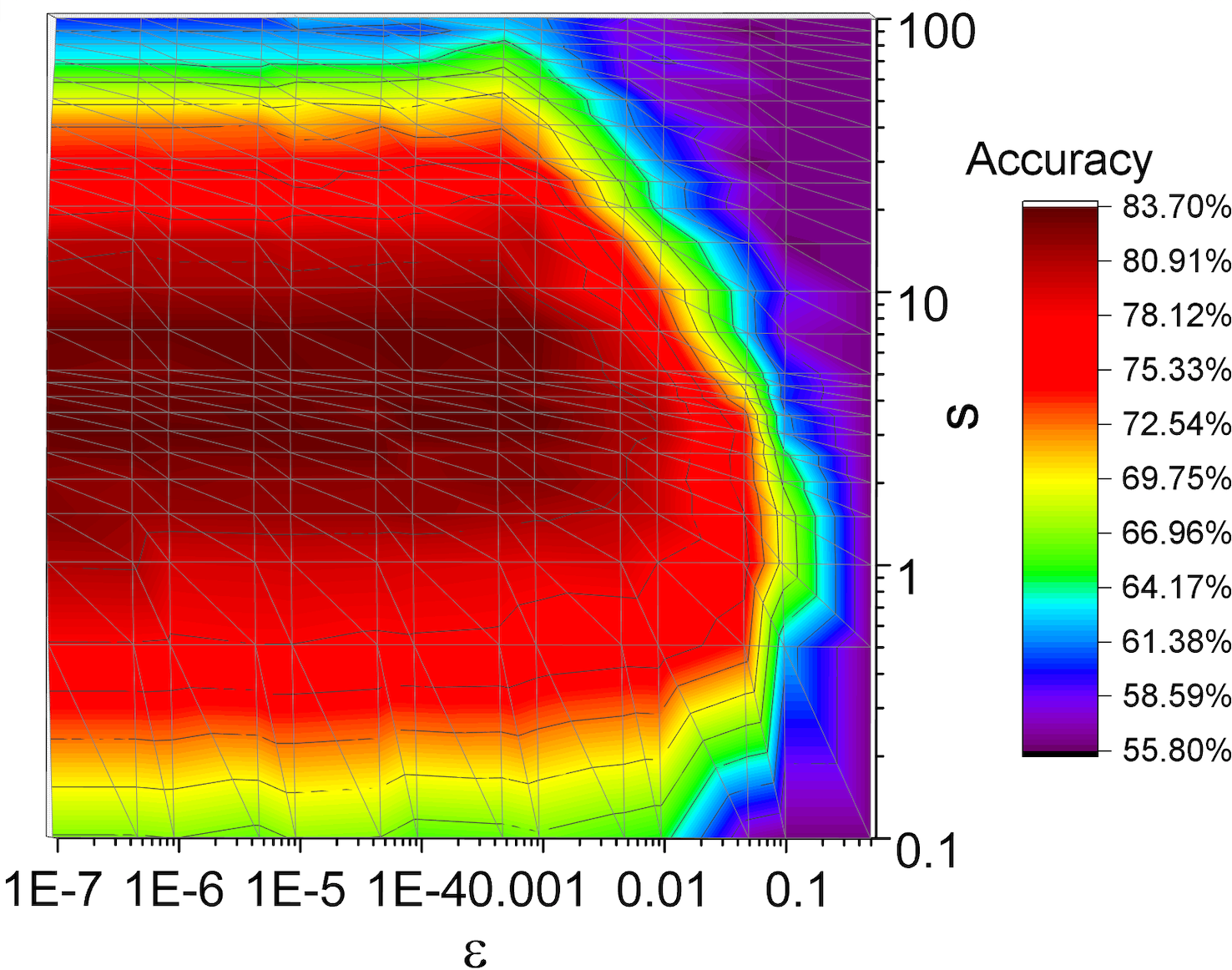}
\caption{The influence of hyper-parameter $s$ and $\epsilon$ on classification accuracy.}
\label{fig:accuracywithparameter}
\end{figure}

Figure~\ref{fig:accuracywithparameter} demonstrates the influence of hyper-parameter $s$ and $\epsilon$ on classification accuracy of Cora. As $s$ becomes larger, the range of neighboring nodes becomes larger, capturing much more information. Figure~\ref{fig:accuracywithparameter} shows that the classification accuracy exactly increases as $s$ becomes larger at first. However, with continuous increase of $s$, some irrelevant nodes are leveraged to update the target node's feature, which violate the smoothness of graph and lead to a drop on accuracy. The classification accuracy decreases a lot with the increasement of $\epsilon$. Using a small $\epsilon$ as threshold can speed up computation and remove noise. However, as $\epsilon$ becomes large, the graph structure is overlooked and some relevant nodes are discarded, then the classification accuracy decreases.

\subsection{Case Study}
\label{Experiments:Case}

\begin{figure}[htbp]
\centering
\includegraphics[width=.28\textwidth]{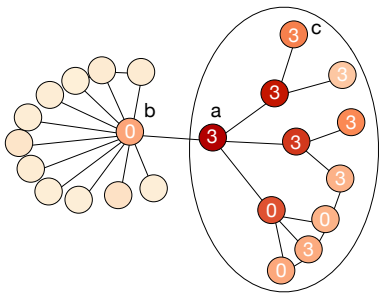}
\caption{Ego network of node a, including first-order and second-order neighbors. Value of each node represents its label. Color of each node represents the strength of signal diffused from node $a$ following heat diffusion. The deeper the color, the larger the value.}
\label{fig:case}
\end{figure}

We conduct case study to illustrate the strengths of our method over competing method, i.e., GCN. Specifically, we select one node in Cora from the set of nodes that are correctly classified by our method and incorrectly classified by GCN. Figure~\ref{fig:case} depicts the two-order ego network of the target node, marked as node $a$. 

GCN leverages the information of the first-order neighbors of node $a$ to assign label to node $a$. Checking the labels of these neighboring nodes, we can find that this is a challenging task for GCN since two neighboring nodes have label $0$ and the other two have label $3$. 

Actually, the smoothness does not depend on order solely. GraphHeat leverages heat kernel to capture smoothness and determine neighboring nodes. When applying our GraphHeat to this case, the neighboring nodes are obtained according to heat diffusion. As shown in Figure~\ref{fig:case}, some second-order neighbors are used to update target node included in the black circle. Moreover, the weight of second-order neighbors can be greater than first-order neighbor, e.g., $[e^{-sL}]_{a,c}>[e^{-sL}]_{a,b}$. These second-order neighbors can enhance our method. At a glimpse of these nodes, we can easily learn why  GraphHeat correctly assign label $3$ to node $a$. In order to capture smoothness, our method enhances the low-frequency filters and discounts high-frequency filters. Instead of depending on order solely, our method exploits  the local structure of target node via heat diffusion. A dense connected local structure can represent strong correlation among nodes. In contrast, although some nodes with high-degree are lower-order neighbors to target node, the connection to target node may represent its popularity rather than correlation, which suggests the weak correlation.

\begin{table}[htbp]
\small
\centering
\setlength{\tabcolsep}{1.42mm}
{
\begin{tabular}{l r r r r r r r}
\hline
\textbf{Index of taget node} & \textbf{12} & \textbf{75} & \textbf{26} & \textbf{1284} & \textbf{1351}  & \textbf{1385} & \textbf{1666}\\
\hline
Neighboring range  & 1 & 2 & 3 & 4 & 5  & 6 & 7\\
Maximum range  & 1 & 2 & 3 & 13 & 13  & 11 & 11\\
\hline
\end{tabular}}
\caption{Varying range of neighboring nodes in GraphHeat}
\label{table:farthestneighbor}
\vspace{0mm}
\end{table}

Finally, we use the Cora dataset to illustrate that the range of neighboring nodes in GraphHeat varies across target nodes. Given a scaling parameter $s$, for some nodes, only a part of their immediate neighbors are included as neighboring nodes, while some nodes can reach a large range of neighboring nodes. To offer an intuitive understanding about the varying range of neighboring nodes, for a target node $a$, we compare the range of neighboring nodes and the potential maximum range, defined as


\vspace{0mm}
\begin{equation}
 \max d_G(a,b) \qquad \forall \  b \in V,
\label{eq13}
\vspace{0mm}
\end{equation}
where $d_G(a,b)$ is the shortest path distance between node $a$ and node $b$. Table 3 shows the results over some randomly-selected target nodes, confirming the flexibility of GraphHeat at defining various range of neighboring nodes. This flexibility is desired in an array of applications~\cite{xu2018representation}.


\section{Conclusion}

The key to graph-based semi-supervised learning is capturing smoothness over graph, however, eigenvectors associated with higher eigenvalues are unsmooth. Our method leverages the heat kernel applied to eigenvalues. This method can discount high-frequency filters and assign larger importance to low-frequency filters, enforcing smoothness on graph. Moreover, we simplify our model to adapt to semi-supervised learning. Instead of restricting the range of neighboring nodes to depend on order solely, our method uses heat diffusion to determine neighboring nodes. The superiority of GraphHeat is shown in three benchmarks and it outperforms than previous methods consistently.

\section*{Acknowledgements}

This work is funded by the National Natural Science Foundation of China under grant numbers
61433014, 61425016 and 91746301. This work is also partly funded by Beijing NSF (No. 4172059). Huawei Shen is also funded by K.C. Wong Education Foundation.

\bibliographystyle{named}
\bibliography{ijcai19}

\end{document}